# Extracting all Aspect-polarity Pairs Jointly in a Text with Relation Extraction Approach

Lingmei BU, Li CHEN, Yongmei LU and Zhonghua YU

Department of Computer Science, Sichuan University, China

**Abstract.** Extracting aspect-polarity pairs from texts is an important task of fine-grained sentiment analysis. While the existing approaches to this task have gained many progresses, they are limited at capturing relationships among aspect-polarity pairs in a text, thus degrading the extraction performance. Moreover, the existing state-of-the-art approaches, namely token-based sequence tagging and span-based classification, have their own defects such as polarity inconsistency resulted from separately tagging tokens in the former and the heterogeneous categorization in the latter where aspect-related and polarity-related labels are mixed. In order to remedy the above defects, inspiring from the recent advancements in relation extraction, we propose to generate aspect-polarity pairs directly from a text with relation extraction technology, regarding aspect-pairs as unary relations where aspects are entities and the corresponding polarities are relations. Based on the perspective, we present a position- and aspect-aware sequence2sequence model for joint extraction of aspect-polarity pairs. The model is characterized with its ability to capture not only relationships among aspect-polarity pairs in a text through the sequence decoding, but also correlations between an aspect and its polarity through the position- and aspect-aware attentions. The experiments performed on three benchmark datasets demonstrate that our model outperforms the existing state-of-the-art approaches, making significant improvement over them.

## 1 Introduction

As a crucial task for fine-grained sentiment analysis, Aspect-Based Sentiment Analysis (ABSA) has attracted considerable attention recently due to its ability to extract aspect mentions and determine their corresponding sentiment polarities expressed in a text, which has more extensive applications than document- and sentence-level sentiment classifications. For instance, given two sentences in Figure. 1, the task of ABSA is to generate 2-tuples (Pizza, positive) and (garlic knots, positive) from Sentence 1, and (Windows 7, positive) and (Vista, negative) from Sentence 2, which denote the commented targets (aspects) and the corresponding sentiment polarities expressed in the sentences, while document- and sentence-level sentiment classifications return only the sentiment polarities as the opinions for these sentences without the commented targets.

| Sentence1:<br>    **Pizza** and **garlic knots** are great as well.<br>APPs:<br>        ( Pizza, positive)  (garlic knots, positive) | Sentence2:<br>    I love **Windows 7** which is a vast improvement over **Vista**.<br>APPs:<br>        (Windows 7, positive)   (Vista, negative) |
|---|---|

**Figure. 1.** Example of Aspect-based sentiment analysis.

Early efforts on ABSA perform extracting aspects and determining their corresponding sentiment polarities separately[1–8], ignoring their interactions and thus gaining worse effectiveness. Recent studies appreciate the interactions and propose a series of joint models to solve the ABSA task, but almost all the models couple the two subtasks through sharing their encoders while the classifiers for different subtasks remain separate, which suffer from the issue of error propagation. Furthermore, the separate classifier approaches are weak at capturing correlations between an aspect and its polarity in a text, further hindering the performance raising.

In order to overcome the shortcomings of the separate classifier approaches, the state-of-the-art solutions adopt token-level sequence tagging and span-level joint classifications. In the token-level tagging approaches[9–11], two different schemas are proposed for aspect-related token BIO tagging and token sentiment polarity tagging, respectively, and sequence2sequence tagging models are adopted for the tagging. The approaches are capable of capturing interactions among aspects and sentiments existing in a text. However, their separate polarity tagging for distinct tokens tend to inconsistent polarities in an aspect. For instance, for the aspect *garlic knots* in Figure 1 composed of two tokens, the polarity sequence tagger could assign different sentiment polarities on these tokens such as *positive* on *garlic* and *negative* on *knots*. The span-level approaches[12–14] mitigate the polarity tagging inconsistency by determining the sentiment polarity towards a span instead of a token. Particularly, they extract spans as candidate aspects and then perform classifications from {*TPOS, TNEG, TNEU, O*} for these spans, where *TPOS, TNET* and *TNEU* denote an aspect with positive, negative, and neutral polarities respectively, and *O* means the span is not a valid aspect at all. However, the categorization schema has a heterogeneous nature where aspect-related and polarity-related labels are mixed，which leads to the difficulty in capturing appropriate knowledge for the two related, but distinct subtasks. Furthermore, although the approaches mentioned above model interactions between aspects and their corresponding sentiment polarities, they overlook the correlations among different aspect-polarity pairs in a text which are critical for accurate sentiment inference of all aspects in a text.

For the purpose of capturing correlations among all the aspect-polarity pairs in a text and accordingly raising performance of ABSA, inspiring by the recent advancements in relation extraction technology[15–17], in this paper we propose to leverage Relation Extraction approach to generate all Aspect-Polarity Pairs(REAPP) directly from a text, considering the task as unary relation extraction where aspects are entities and the corresponding polarities are the relations. Based on the perspective, we present a sequence2sequence model to capture not only relationships among aspect-polarity pairs in a text but also correlations between an aspect and its polarity through a position- and aspect-aware attentions for extracting aspect-polarity pairs in a text jointly. We conduct a series of experiments on three well-known benchmark datasets to verify the effectiveness of the model. The experimental results demonstrate that our model indeed significantly improves the performance, verifying benefits from incorporating correlations among aspect-polarity pairs in a text for their joint recognition and extraction.

The major contributions of this work are three-fold: (1) the task of Aspect-Based Sentiment Analysis (ABSA) is boiled down to an unary relation extraction problem in order to capture interactions among different aspect-polarity pairs in a text and overcome the shortcomings existing in the state-of-the-art approaches, namely polarity inconsistency in token tagging and heterogeneous nature in span-level classifications; (2) a position- and aspect-aware sequence2sequence model for extracting aspect-polarity pairs in a text jointly is proposed; (3) extensive experiments are conducted to investigate the performance of our model for ABSA, and the experimental results verify its effectiveness.

## 2     Related work

Sentiment analysis aims to identify attitudes expressed in a text by its author(s) and has drawn much attention in the domain of natural language processing due to its wide applicability. However, early efforts on sentiment analysis have focused mostly on determining sentiment polarities in a document or a sentence such as positive, negative or neutral without the targets commented[18,19,20], which has only limited ability to support making public-opinion-based decisions. Therefore, majority of recent related works turn to fine-grained sentiment analysis which involves identifying target aspects and classifying their corresponding sentiments in a text.

At first the two subtasks in fine-grained sentiment analysis are investigated separately, and a series of approaches are proposed for target aspect identification[1,2,3] and sentiment classification towards given annotated aspects[4,5,6,8,32] respectively. However, it is obvious that separate aspect-oriented sentiment classification is limited in its applicability. Pipelining the two subtasks, of course, is able to perform fine-grained sentiment analysis without the requirement of specifying target aspects in advance, but the pipelined process suffers from the issue of error propagations, degrading the effectiveness in fine-grained sentiment analysis. Solving the two subtasks simultaneously and jointly thus becomes a current hotspot in the research area.

All the joint approaches existing nowadays to extract target aspects and determine their corresponding sentiment polarities simultaneously in a text are divide into two groups. The first group solves the task by means of sequence labeling[9][10][11], where a tag has mixed an original aspect-related tag from {B-Begin, I-Inside, O-Other} with a sentiment polarity from {positive, negative, neutral}. Due to the mixing nature of the tags, how to capture the two types of information and integrate them effectively to predict a mixed tag for a token accurately becomes a key factor for the success of the sequence labeling. The study of Mitchell et al.[11] is the first such attempt. They employ CRF model for the sequence labeling with the collapsed tags and compare it with the other two trivial sequence labeling approaches, namely sequence labeling aspects and sentiment polarities separately, i.e., completely pipelined, and the two sequence labeling processes coupled together with the hidden aspect tags being features for the polarity tagging. But unfortunately, the experimental results demonstrate that the former does not gain consistent improvement over the latter two. Zhang et al.[10] incorporate the idea of Mitchell et al.[11] into a neural network model and further consider the constraints among different tags for a token. They likewise do not verify the superiority of the former, complete end-to-end approach over the latter two approaches. Li et al.[9] further strengthen the end-to-end sequencing labeling with the collapsed tags, taking the previous token embedding into consideration for tagging a token in order to tackle the issue of sentiment inconsistency, i.e., distinct sentiment polarities are assigned to the tokens in an aspect. However, the strengthened method is unable to avoid the inconsistent issue.

The difficulties in the existing sequence labeling approaches, particularly the issue of sentiment inconsistency inherent to the token-level sentiment tagging for larger unit sentiment analysis, i.e., sentiment analysis towards aspects often containing several tokens, stimulate the appearance of the second group of approaches, namely span-based approaches. In this vein Hu et al.[12] propose to leverage two span-related classifiers for recognizing span start tokens and end tokens, and then determine sentiment polarities of the identified spans. Except sharing parameters and encodings, the two classifiers operate independently, which results in the necessity of additional heuristic step for pairing the start and end of a valid span. Besides, the approach of Hu et al.[12] determines sentiment polarity of a span depending only on the contextual embeddings of tokens in the span, ignoring the tokens out of the span. However, it is obvious that the contextual tokens out of the span are bearing the sentiment polarity of the span. Another similar work is provided by Lin et al.[14]. They improve the approach of Hu et al.[12] from two aspects. The first one is to leverage two multi-binary classifiers for determining start tokens and end tokens of spans, which is more reasonable mathematically. The second one is a more quality heuristic step for pairing the start and end of a valid span. The improvements bring superiority in effusiveness over the original counterpart, however, the issues intrinsic to the approach Hu et al.[12] remain in the approach of Lin et al.[14].

Zhou et al.[13] integrate span(target aspect) extraction and sentiment polarity determination into one classification step. They first generate all possible spans up to a certain length in a sentence, and then invoke a multiclass classifier to determine the class of a span from {TPOS, TNEG, TNEU, O}, where the first three class labels denote being target aspect with positive, negative and neutral polarities, respectively, and the last one means other. Obviously, the set consists of heterogeneous classification labels. Such heterogeneity in classification scheme could hinder the classification performance. Furthermore, the approach of Zhou et al.[13] classify all candidate spans in a sentence independently, ignoring their interactions which are critical because the sentiment polarity of a particular aspect inevitably has influence on the sentiment polarities of the other aspects in the same sentence.

In order to overcome the issues of the existing approaches, inspired by recent advances in relation extraction technology, we propose to treat the joint task of extracting aspects and degerming their corresponding sentiment polarities in a sentence as an end-to-end relation extraction process with the entities given or recognized in advance, where the entities are aspects and the relations are sentiment polarities. Because we have only one entity for a relation, the relation to be extracted is in fact a unary relation. Similar to our task, relation extraction also has undergone a pipelined stage[23–26], and a joint stage[15–17] appearing just recently. Zeng et al.[15] and Nayak et al.[17] offer latest joint methods with which all the relational triplets for a sentence could be generated simultaneously. The ideas in these methods are appropriate for our task. However, the information to be extracted in these methods is binary relations, while our task is to extract unary relations. In addition, our relation classes are sentiment polarities, which are expressed by adjectives in general cases, but for the relation extraction, the relations are always expressed by verbs.

Another related task is extracting aspect-opinion mention pairs in texts without the polarity determination[22], whereas in our task sentiment polarity towards a target aspect is the dominating concern.

## 3    Proposed Model

In order to solve the issues in the existing approaches to ABSA task, we propose to regard the task as a relation extraction process where the aspects to be extracted are entities and their corresponding sentiment polarities are the relations. It is obvious that the relations are unary, but if we represent an aspect with its span in the sentence, the relations become triplets similar to relations in typical information extraction tasks. In this section we elaborate on a position- and aspect-aware sequence2sequence model devised specially for the ABSA task from our relation extraction perspective.

### 3.1    Task Description

Given a sentence $S = \{w_1, w_2, \cdots, w_n\}$ consisting of $n$ tokens $w_1, w_2, \cdots, w_n$, our task is to generate a set of triplets $Y = \{(s, e, r) | 0 \leq s \leq e < n, r \in R\}$, where $s$ and $e$ represent the start and end positions of an aspect in $S$ respectively, and $r$ is the corresponding sentiment polarity of the aspect expressed in $S$. The set $R$ is composed of all possible sentiment polarities, i.e., $R = \{'POS', 'NEG', 'NEU'\}$ with its components meaning positive, negative, and neutral polarities respectively. For instance, for the second sentence in Figure 1, our task is to generate triplets $(2, 3, 'POS')$ and $(10, 10, 'NEG')$ from it, which denote aspect-pairs (Windows 7, positive) and (Vista, negative) respectively, as illustrated in Table 1.

**Table 1.** An example for REAPP

| | | |
|---|---|---|
| *Sentence* | $I_0$ love$_1$ Windows$_2$ 7$_3$ which$_4$ is$_5$ a$_6$ vast$_7$ improvement$_8$ over$_9$ Vista$_{10}$ .$_{11}$ | |
| *Triplets* | (2, 3, POS) | (10, 10, NEG) |
| *Aspect-polarity pair* | (Windows 7, positive) | (Vista, negative) |

## 3.2 The Model Architecture

The architecture of our position- and aspect-aware sequence2sequence is composed of 4 layers, as depicted in Figure 2. The first layer is embedding layer, which is responsible for initializing all the tokens in a sentence with their corresponding word embeddings. The second layer, encoding layer, performs sentence analysis based on the initialized word embeddings in order to capture interactions among all the tokens in the sentence and accordingly learn refined token contextual representations suited for subsequent aspect span and sentiment polarity generations. The last two layers, i.e., the position- and aspect-aware attention layer and decoding layer, are tightly coupled. The attention layer provides different kinds of context computations for the decoding layer, depending on distinct decoding temporal steps such as the first step for decoding a span start position and the third step for decoding a sentiment polarity. The computations are performed with a position- and aspect-aware attention mechanism. The details are exposed in subsection 3.3. The decoding layer is devoted to generating an argument of a triplet during every temporal step, taking a contextual representation suited for the current temporal step as the input. It is noticed that, the label $NA$ is introduced to denote the boundary of an input sentence and as well to note the termination of a decoding process. The decoding layer is detailed in subsection 3.4.

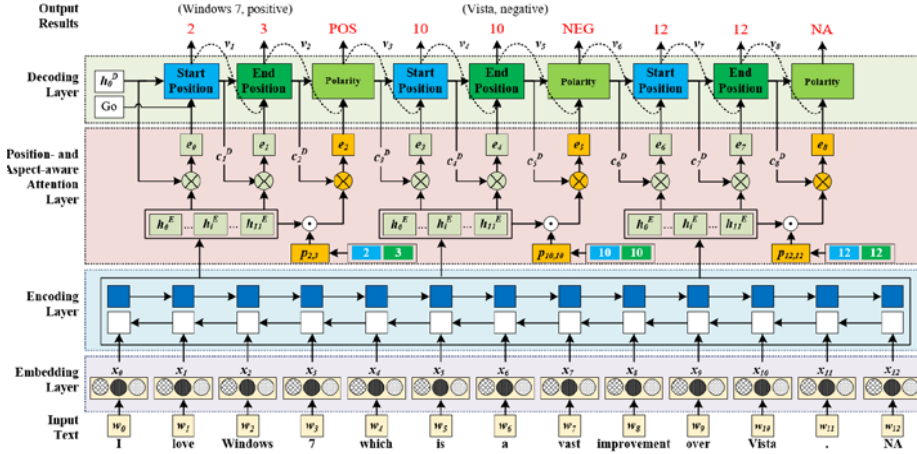

Figure 2. Architecture of the proposed model.

We employ the pre-trained GloVe word vectors from [文献] to initialize token representations in an input sentence. Furthermore, in order to cope with the issue of OOV (Out of Vocabulary) and considering the roles of POS (Part of Speech) in determining aspects and sentiments such as aspects always being noun phrases and sentiments always expressed with adjectives, we incorporate POS embeddings and char-level embedding into the initialized representations. Concretely, for a token $w_i$, we encode it with convolutional operations on its character sequence followed by max-pooling[31], and such resulted encoding $e_i^c \in R^{d_f}$ is concatenated with the pre-trained GloVe vector $e_i^w \in R^{d_w}$ of the token and its POS embedding $e_i^p \in R^{d_p}$ to initialize the token representation $x_i$, i.e., $x_i = [e_i^w; e_i^c; e_i^p]$. After that we employ Bi-LSTM to analyze the whole sentence and further capture the contextual information for a token. We concatenate the forward and backward hidden states of a token as its contextual embedding, i.e., for $w_i$, its contextual embedding is

$$h_i^E = [\overrightarrow{LSTM}(x_i, h_{i-1}^E); \overleftarrow{LSTM}(x_i, h_{i-1}^E)] \quad (1)$$

### 3.3 Position- and Aspect-aware Attention Layer

This layer prepares contextual representations of an input sentence for different temporal steps in the sequence decoding process, thereby being closely related to the decoding layer. Since the task of the decoding layer is to generate a triplet sequence where each triplet is composed sequentially of an aspect span (an aspect start position followed by its end position) and the corresponding sentiment polarity, i.e., the decoded output is structured, which kind of arguments in a relational triplet is to be generated for a particular temporal step is predictable. Concretely, suppose the current decoding step is t, t % 3 = 1 means the kind of arguments to be generated in this temporal step is an aspect start position, while t % 3 = 2 means the kind of arguments is the end position of the aspect whose start position is the decoded result at step t – 1, and t % 3 = 0 denotes the kind of arguments is the sentiment polarity of the aspect generated at steps t – 2 and t – 1. Various kinds of arguments require diverse contextual information of the input sentence. Therefore, we provide different contextual encodings of the sentence with position- and aspect-aware attentions in accordance with different values of t. The details are exposed below.

For the temporal steps when t % 3 = 1 or t % 3 = 2, we employ the attention mechanism in [32] to compute the contextual representation $e_t$ of the input sentence, as shown in the formulas (2), (3) and (4), where n is the length of the input sentence, $c_{t-1}^D$ is the state vector of the decoding step t – 1 transferred to the current step t, and $h_i^E$ is the contextual embedding for the token $w_i$ obtained through formula (1). The symbols $W^P$, $b^P$, $W^Q$ and $v^P$ are learned parameters of our model.

$$m_t = W^P c_{t-1}^D + b^P, \quad n_{t,i} = W^Q h_i^E, \quad (2)$$

$$\beta_{t,i} = v^P \tanh(m_t + n_{t,i}), \quad \alpha_t^P = softmax(\beta_t), \quad (3)$$

$$e_t = \sum_{i=0}^{n-1} \alpha_{t,i}^P \times h_i^E, \quad (4)$$

Even though the contributions of tokens in a sentence are learned here regardless of t % 3 = 1 and t % 3 = 2, we distinguish them with different mask strategies, with which a token indirectly contribute its evidence differently to the two distinct decision tasks, namely determining an aspect start position and determining the corresponding end position. The mask strategies are detailed in subsection 3.4.

For the temporal steps when t % 3 = 0, we compute the contextual representation for step t differently due to the particular task of the decoder. In such steps the decoder should generate an appropriate sentiment polarity towards the aspect recognized at the immediately preceding two steps. It is obvious that the tokens nearby the current aspect are more important than the tokens far away from the aspect. Therefore, we take the distance of a token to the current aspect into consideration to determine the contribution of the token with regard to determination of the sentiment polarity towards the aspect. We first scale $h_i^E$ according to the distance of $w_i$ to the current aspect, and then determine the attention of $w_i$ for the contextual representation with similar manner as for steps when t % 3 = 1 and t % 3 = 2. The contextual representation is obtained with the formulas (5) – (8) below.

$$p_{s,e}^i = 1 - \frac{d_i}{n}, \tag{5}$$

$$m_t = W^A c_{t-1}^D + b^A, \; n_{t,i} = W^R \left( p_{s,e}^i h_i^E \right), \tag{6}$$

$$\beta_{t,i} = v^A \tanh(m_t + n_{t,i}), \; \alpha_t^A = softmax(\beta_t), \tag{7}$$

$$e_t = \sum_{i=0}^{n-1} \alpha_{t,i}^A (p_{s,e}^i h_i^E), \tag{8}$$

where $d_i$ is the distance between $w_i$ and the aspect identified at the steps t – 2 and t – 1, $W^A$, $b^A$, $W^R$ and $v^A$ are learned parameters of our model, and the other notations are the same as in formulas (2) – (4).

Except the tokens nearby an aspect, the aspect itself is also an important cue for determining its sentiment polarities expressed in the sentence. For this reason, in the decoding phrase we will employ a Bi-LSTM to encode the aspect and the encoding is integrated with $e_t$ to perform the polarity determination. The details are described in subsection 3.4.

### 3.4 Decoding Layer

In this layer we utilize a LSTM to generate the triplet sequence for an input sentence. To this end, for a temporal step t, the LSTM decoder first fuse the contextual representation of the sentence for this step with the information transferred from the step t – 1 to obtain its current states $h_t^D$ and $c_t^D$, shown in formula (9), and then generates the argument of the current triplet for this temporal step depending on different value of t. The different operations for different t are detailed subsequently. The notation $v_{t-1}$ denotes the embedding of the triplet argument generated at t – 1. All possible decoded triplet arguments such as single tokens (corresponding to the generated *e* and *s*), *NA* and the sentiment polarities from *R* are composed into our decoding vocabulary. The embeddings for this vocabulary are initialized randomly and learned jointly with the other learned parameters of our model.

$$h_t^D, c_t^D = LSTM([v_{t-1}; e_t] \cdot W^u, c_{t-1}^D), \tag{9}$$

During the step of $t\%3 = 1$, the decoder should generate a probabilistic distribution over all the tokens in the input sentence meaning the start of the current aspect to be generated, and *NA* meaning termination of the whole generation process for the sentence. To this end, we concatenate $h_t^D$ with every token's contextual representation $h_i^E$ and pass the concatenation through a Feed-Forward Network (FFN) to compute the probability of the token. The probability of NA is computed with a slightly different manner. We simply pass $h_t^D$ into another FFN to determine the situation where generating the whole triplet sequence is completed. Besides the operations introduced above, in order to remedy the issue of duplicate generation, we incorporate a mask mechanism with which we mask the positions covered by the aspects generated before for the sentence so that a token could appear in at most one aspect. The concrete computations in this temporal step are described in formulas (10) – (13), where $W^s$, $W^{NA}$, $b^s$ and $b^{NA}$ are all learned parameters of our model, $M^s \in R^n$ are mask vector, E is all the aspects generated before t for the sentence, and $p^s$ is the probabilistic distribution for the current step.

$$q_i^s = selu([h_t^D; h_i^E] \cdot W^s + b^s), \quad (10)$$

$$q^{NA} = selu(h_t^D \cdot W^{NA} + b^{NA}), \quad (11)$$

$$M_i^s = \begin{cases} 1, i \in E \\ 0, i \notin E \end{cases} \quad (12)$$

$$p^s = softmax([M^s \otimes q^s; q^{NA}]), \quad (13)$$

For the step t when $t\%3 = 2$, the decoder needs to determine the aspect end position whose start position has been determined at the step t – 1. For this task we incorporate another mask vector $M^e \in R^n$ to ensure the end position appearing after the start position in the sentence. The other operations are similar as those for $t\%3 = 2$. We summarize the process for $t\%3 = 2$ with formulas (14) – (16), where $s$ is the start position determined at the step t – 1, $W^e$ and $b^e$ are learned parameters, $q^{NA}$ is computed with formula (11) based on the current step's $h_t^D$, and $p^e$ is the probability distribution of the tokens being the end position of the aspect.

$$M_i^e = \begin{cases} 1, i \geq s \\ 0, i < s \end{cases} \quad (14)$$

$$q_i^e = selu([h_t^D; h_i^E] \cdot W^e + b^e), \quad (15)$$

$$p^e = softmax([M^s \otimes q^e \otimes M^e; q^{NA}]). \quad (16)$$

During the step t when $t\%3 = 0$, the task of the decoder is to predict the sentiment polarity for the aspect identified at the preceding two steps. Except its context, the aspect itself is also an important cue for determining the sentiment polarity towards it. Therefore, we employ a BiLSTM on the token sequence $h_s^E, h_{s+1}^E, \cdots h_e^E$ of the aspect, and encode the aspect with $h_a = [h_s^E; h_{sec}; h_e^E]$, where $h_{sec}$ is the concatenation of the two final hidden vectors obtained in the forward and backward processes respectively. After that the similar processing as for the other two situations is performed to generated a probabilistic distribution over the three possible sentiment polarities and *NA*. The details are expressed I formulas (17) – (19), where $W^r$, $b^r$, $W^{SNA}$ and $b^{SNA}$ are learned parameters, and $p^r$ is the probability distribution over all the three possible sentiment polarities and *NA*.

$$q_j^r = selu([h_a; h_t^D] \cdot W^r + b^r), j \in [1, 3] \quad (17)$$

$$q^{SNA} = selu(h_t^D \cdot W^{SNA} + b^{SNA}) \quad (18)$$

$$p^r = softmax([q^r; q^{SNA}]), \tag{19}$$

We train our model with negative log-likelihood loss to be minimized. Suppose the training set is $X = \{< S_1, Y_1 >, < S_2, Y_2 >, \cdots, < S_B, Y_B >\}$, where $S_i$ is a sentence, $Y_i$ is its ground truth annotation with $Y_i = \{y_i^1, y_i^2, \cdots y_i^T\}$ denoting triplet arguments for the sentence $S_i$, then Loss is calculated as

$$\mathcal{L}_{loss} = -\frac{1}{B \times T} \Sigma_{i=1}^{B} \Sigma_{t=1}^{T} \log(p(y_i^t | y_i^{<t}, x_i)). \tag{20}$$

## 4   Experiments

### 4.1   Datasets

We validate the effectiveness of our model on three benchmark datasets widely accepted for the task of ABSA, namely LAPTOP[32], REST[32–34] and TWITTER[11]. The statistics and the separations of training, development and testing data are shown in Table 2. For ease of comparisons, we choose the same separations of training, development and testing data to perform out experiments on LAPTOP. For TWITTER without the separation, we experiment our model on it with 10-fold cross-validation fashion, as done in [11]. The evaluation metrics used here are the same as in the related studies, including precision, recall, $F_1$-value and accuracy. Furthermore, a predicted aspect is counted as a correct one only if it exactly matches with the corresponding aspect ground truth annotation.

**Table 2.** Statistics of the three datasets

| Datasets | LAPTOP | | | REST | | | TWITTER | | |
|---|---|---|---|---|---|---|---|---|---|
| | POS | NEG | NEU | POS | NEG | NEU | POS | NEG | NEU |
| training | 883 | 754 | 404 | 2337 | 942 | 614 | | | |
| dev | 104 | 106 | 46 | 270 | 93 | 50 | - | - | - |
| testing | 339 | 130 | 165 | 1524 | 500 | 263 | | | |
| total | 1326 | 990 | 615 | 4131 | 1535 | 927 | 692 | 263 | 2244 |

### 4.2   Experiment Settings

We initialize embeddings for the words in our encoding vocabulary with the glove.840B.300d[1][30], and these embeddings are further trained jointly with the other parameters of our model, while the word embeddings in the decoding vocabulary are initialized randomly and learned during the training process of our model. We invoke the Stanford CoreNLP toolkit[2] to perform POS tagging for tokens in a sentence, and the dimensionality of a POS tag embedding is 50. For the character-level token encoding with CNN and max-polling we utilize 50 filters of size 3, thus resulting in a token character-level embedding of dimensionality 50. The contextual token encodings for an input sentence are learned by using two stacked BiLSTMs, where forward and backward dimensionalities are all 150-dimensional, therefore the resulted contextual token encodings are 300-dimensional. For the decoding LSTM, we also set its dimensionality to 300. We employ the Adam optimizer with the learning rate of 0.001 and dropout probability of 0.5 to train our model, and the batch size is 32.

---

[1] https://nlp.stanford.edu/projects/glove/
[2] https://stanfordnlp.github.io/CoreNLP/

### 4.3 Baselines

The baselines for our empirical comparisons include all the state-of-the-art approaches and their variations proposed recently. Concretely, the baselines are:

- CRF-{pipelined, joint, collapsed}(2013)[11]: a CRF sequence-labeling method based on an emotion lexicon and manually-built features;
- NN-CRF-{pipelined, joint, collapsed}(2015)[10]: a shallow neural network model with CRF sequence-labeling process on its top to capture correlations between labels;
- UNIFIED[9](2019): a BiLSTM-based sequence-labeling method with a collapsed tag set;
- TAG- {pipelined, joint, collapsed}[12]: a CRF-based sequence-labeling model with BERT as its encoder;
- Zhou-SPAN(2019)[13]: a span-based joint method employing a span-aware attention mechanism to get the sentiment information for each span;
- Hu-SPAN-{pipelined, joint, collapsed}(2019)[12]: a span-based method with BERT as its encoder and a heuristic algorithm for pairing aspect start positions with their corresponding end positions;
- SPRM[14](2020): an improvement over Hu-SPAN-joint[12] with a private representation for every single task(i.e., aspect identification task and sentiment polarity prediction task) and a share representation for the joint task to capture correlations between the two tasks.

### 4.4 Experimental Results

The experimental results for aspect-polarity extraction on the three benchmark datasets are listed in Table 3. It is obvious from the empirical data in Table 3 that our model consistently superior to the state-of-the-art methods, regardless of whether they are sequence-labeling-based or span-based. The only exception is the recall on REST which is slightly lower than that of SPRM (2020), but the precision and F1-values are increased by 4.23% and 1.96%, respectively.

In order to verify roles different components of our model act for the task, we further perform a series of ablation studies, and the results are listed in the last five rows of Table 3. Comparing these results, we can find that the character-level token embedding with CNN followed by max-pooling is not significant for our model, which only leads to a tiny improvement in the task performance. In contrast, incorporating aspect embedding and aspect-aware attention mechanism could provide more gains to our model for solving the task.

Table 3. Experimental results for aspect-polarity pair extraction on three datasets

| | Method | LAPTOP | | | REST | | | TWITTER | | |
|---|---|---|---|---|---|---|---|---|---|---|
| | | P(%) | R(%) | F1(%) | P(%) | R(%) | F1(%) | P(%) | R(%) | F1(%) |
| sequence labeling methods | CRF-pipeline (2013) | 59.69 | 47.54 | 52.93 | 52.28 | 51.01 | 51.64 | 42.97 | 25.21 | 31.73 |
| | CRF-joint (2013) | 57.38 | 35.76 | 44.06 | 60.00 | 48.57 | 53.68 | 43.09 | 24.67 | 31.35 |
| | CRF-collapsed (2013) | 59.27 | 41.86 | 49.06 | 63.39 | 57.74 | 60.43 | 48.35 | 19.64 | 27.86 |
| | NN-CRF-pipeline (2015) | 57.72 | 49.32 | 53.19 | 60.09 | 61.93 | 61.00 | 43.71 | 37.12 | 40.06 |
| | NN-CRF-joint (2015) | 55.64 | 34.48 | 45.49 | 61.56 | 50.00 | 55.18 | 44.62 | 35.84 | 39.67 |
| | NN-CRF-collapsed (2015) | 58.72 | 45.96 | 51.56 | 62.61 | 60.53 | 61.56 | 46.32 | 32.84 | 38.36 |
| | UNIFIED (2019) | 61.27 | 54.89 | 57.90 | 68.64 | 71.01 | 69.80 | 53.08 | 43.56 | 48.01 |
| | TAG-pipeline (2019) | 65.84 | 67.19 | 66.51 | 71.66 | 76.45 | 73.98 | 54.24 | 54.37 | 54.26 |
| | TAG-joint (2019) | 65.43 | 66.56 | 65.99 | 71.47 | 75.62 | 73.49 | 54.18 | 54.29 | 54.20 |
| | TAG-collapsed (2019) | 63.71 | 66.83 | 65.23 | 71.05 | 75.84 | 73.35 | 54.05 | 54.25 | 54.12 |
| span-based methods | Zhou-SPAN (2019) | 61.40 | 58.20 | 59.76 | 76.20 | 68.20 | 71.98 | 54.84 | 48.44 | 51.44 |
| | Hu-SPAN-pipeline (2019) | 69.46 | 66.72 | 68.06 | 76.14 | 73.74 | 74.92 | 60.72 | 55.02 | 57.69 |
| | Hu-SPAN-joint (2019) | 67.41 | 61.99 | 64.59 | 72.32 | 72.61 | 72.47 | 57.03 | 52.69 | 54.55 |
| | Hu-SPAN-collapsed (2019) | 50.08 | 47.32 | 48.66 | 63.63 | 53.04 | 57.85 | 51.89 | 45.05 | 48.11 |
| | SPRM (2020) | 68.66 | 68.77 | 68.72 | 77.78 | **80.60** | 79.17 | 60.25 | 58.79 | 59.45 |
| ours | Ours | **70.79** | **70.68** | **70.73** | **82.01** | 80.27 | **81.13** | **63.18** | **59.97** | **61.53** |
| | - POS | 70.32 | 69.69 | 70.00 | 81.24 | 80.12 | 80.68 | 62.26 | 59.34 | 60.76 |
| | - character embedding | 70.62 | 70.52 | 70.57 | 81.56 | 80.02 | 80.78 | 63.06 | 59.58 | 61.27 |
| | - POS and character embeddings | 70.27 | 69.37 | 69.82 | 80.18 | 80.11 | 80.14 | 61.85 | 59.3 | 60.55 |
| | - BiLSTM-based aspect embedding | 69.65 | 68.92 | 69.28 | 80.71 | 79.15 | 79.92 | 61.54 | 59.48 | 60.49 |
| | -aspect-aware attention | 69.82 | 68.78 | 69.30 | 80.99 | 79.18 | 80.07 | 61.72 | 59.37 | 60.52 |

In addition, we investigate the performance of our model for every single subtask, i.e., aspect identification and sentiment polarity prediction respectively, and the results are compared with those of representative state-of-the-art sequence-labeling and span-based approaches, namely TAG-pipeline (2019), Hu-SPAN-pipeline (2019) and SPRM (2020). The results are shown in Table 4. We can see in Table 4 that our model performs every single subtask better that the state-of-the-art approaches, consistently superior to them on the performance measures F1-value and accuracy, except the F1-value of aspect identification for TWITTER, whose highest value is obtained by Hu-SPAN-pipeline (2019).

Table 4. Experimental results for every single subtask on three datasets

| model | Aspect identification (F1%) | | | Polarity prediction (Acc %) | | |
|---|---|---|---|---|---|---|
| | LAPTOP | REST | TWITTER | LAPTOP | REST | TWITTER |
| TAG-pipeline | 85.20 | 84.48 | 73.47 | 71.42 | 81.80 | 59.76 |
| Hu-SPAN-pipeline | 83.35 | 82.38 | **75.28** | 81.39 | 89.95 | 75.16 |
| SPRM | 84.72 | 86.71 | 69.85 | 81.50 | 90.35 | 78.34 |
| Ours | **85.31** | **87.55** | 74.88 | **81.75** | **91.52** | **80.83** |

## 5 Conclusion

In this paper, we propose a position- and aspect-aware sequence2sequence model for joint extraction of aspect-polarity pairs. Different from current token level sequence labeling models and span-based models, our model take advantage the relationships among aspect-polarity pairs in a text and thus is able to capture interactions between two subtasks, namely aspect identification and sentiment polarity prediction. Moreover, the polarity prediction is performed on aspect level instead of token level, thus avoiding the issue of polarity inconsistency inherent in the existing token-level sequence labeling approaches and the issue of heterogeneity among class labels employed in the span-based approaches. The experiments conducted on three benchmark datasets demonstrate that our model clearly and consistently outperforms the existing state-of-the-art approaches.